# Pretrained Domain-Specific Language Model for General Information Retrieval Tasks in the AEC Domain

Zhe Zheng[1], Xin-Zheng Lu[1], Ke-Yin Chen[1], Yu-Cheng Zhou[1], Jia-Rui Lin[1,*]

*Corresponding author, E-mail: lin611@tsinghua.edu.cn; jiarui_lin@foxmail.com

(1. Department of Civil Engineering, Tsinghua University, Beijing, 100084, China)

**Abstract:**

As an essential task for the architecture, engineering, and construction (AEC) industry, information retrieval (IR) from unstructured textual data based on natural language processing (NLP) is gaining increasing attention. Although various deep learning (DL) models for IR tasks have been investigated in the AEC domain, it is still unclear how domain corpora and domain-specific pretrained DL models can improve performance in various IR tasks. To this end, this work systematically explores the impacts of domain corpora and various transfer learning techniques on the performance of DL models for IR tasks and proposes a pretrained domain-specific language model for the AEC domain. First, both in-domain and close-domain corpora are developed. Then, two types of pretrained models, including traditional wording embedding models and BERT-based models, are pretrained based on various domain corpora and transfer learning strategies. Finally, several widely used DL models for IR tasks are further trained and tested based on various configurations and pretrained models. The result shows that domain corpora have opposite effects on traditional word embedding models for text classification and named entity recognition tasks but can further improve the performance of BERT-based models in all tasks. Meanwhile, BERT-based models dramatically outperform traditional methods in all IR tasks, with maximum improvements of 5.4% and 10.1% in the F1 score, respectively. This research contributes to the body of knowledge in two ways: 1) demonstrating the advantages of domain corpora and pretrained DL models and 2) opening the first domain-specific dataset and pretrained language model for the AEC domain, to the best of our knowledge. Thus, this work sheds light on the adoption and application of pretrained models in the AEC domain.

**Keywords:**

information retrieval (IR); text classification (TC); named entity recognition (NER); natural language processing (NLP); transfer learning; pretrained BERT model; automated rule checking (ARC)

# 1 Introduction

The architecture, engineering, and construction (AEC) industry is a significant driver of economic activity around the world. Currently, the AEC industry is undergoing a significant transformation from conventional labor-intensive methods to automation through the adoption of information and automation technologies (Wang et al., 2020; Wu et al., 2022a; Liao et al., 2021). To enhance the automation and informatization of the AEC industry, it is essential to make full use of all kinds of data, including structured data and unstructured data. However, over 80% of data in the AEC sector are unstructured, and most of them are texts, which are difficult to handle (Wu et al., 2022a). Hence, many efforts have been devoted to natural language processing (NLP) and information retrieval (IR) technologies, which aim to extract and understand text data automatically and intelligently. Recently, NLP and IR have been increasingly adopted in the AEC sector and have been applied in many applications, including filtering information, organizing documents, expert systems, and automated rule checking (ARC) (Hassan et al., 2021; Fuchs, 2021; Wu et al., 2022a). Among them, the ARC is one of the main application scenarios (Wu et al., 2022a), which involves almost all typical NLP and IR tasks in the AEC domain. ARC aims to automate the compliance check process of a building design with applicable regulatory texts. The ARC process can be subdivided into four stages: 1) rule interpretation, 2) building model preparation, 3) rule execution, and 4) reporting checking results (Eastman et al., 2009; Ismail et al., 2017). Among the four stages, rule interpretation, which translates regulatory text into a computer-processable format, is the most vital and complex stage needing further investigation for IR (Ismail et al., 2017).

In recent years, with emerging deep learning (DL) techniques and open datasets for model training, it has become possible for deep NLP-based methods to achieve a more comprehensive understanding of regulatory texts (Fuchs, 2021). As a result, many IR methods have been further developed and improved based on DL techniques (Fuchs, 2021). Generally, the deep learning-based IR methods used in AEC can be mainly divided into two tasks:

**(1) Text classification (TC)** divides texts into different groups. Typical applications of TC include a) recognizing relevant sentences in a regulatory text corpus, thereby avoiding inefficiency and errors in downstream tasks (e.g., NER) resulting from unnecessary processing of irrelevant text (Zhou & El-Gohary, 2016; Song et al., 2018), and b) identifying construction site accidents from large quantities of documents (Cheng et al., 2020).

**(2) Named entity recognition (NER)** detects semantic elements in a sentence and is also known as information extraction (IE) in the AEC domain. Typical applications of NER include a) identifying and extracting the words and phrases in the relevant sentences to interpret rules into computer-processable representations (Zhou et al., 2020; Zhang & El-Gohary, 2021; Moon et al., 2021) and b) automatic construction and extensions of knowledge graphs (Leng et al., 2019; Wu et al., 2022b).

Compared to traditional machine learning models, deep learning methods have significantly more parameters and typically need a larger scale of data for training (Zhang & El-Gohary, 2021). However, there is still a lack of unified semantic labels in the AEC domain, and few public training datasets are available for further investigation. As a result, deep learning-based approaches require highly expensive manual effort to prepare sufficient training datasets (Xu & Cai, 2021). Due to the lack of labeled training datasets, an unsupervised learning technique called model pretraining is introduced, which can learn rich syntactic and semantic patterns from large-scale textual datasets collected from the internet. Thus, it is possible to improve the performance of DL methods by transfer learning techniques that lack large-scale labeled datasets. For example, the bidirectional encoder representation from transformers (BERT) proposed by Devlin et al. (2018), a widely used NLP model based on DL, is pretrained on a large-scale corpus with 3,300 M words (Devlin et al., 2018). In this way, the proposed BERT model can learn hidden knowledge from large-scale datasets by pretraining and further help various downstream tasks by transfer learning (Fang et al. 2020). Applications in bioinformatics (Mohan et al., 2021) and ARC (Fang et al., 2020; Zhou et al., 2020) have illustrated the advantages of pretrained language representation models such as BERT.

However, as pointed out by Sun et al. (2019), the widely used BERT models are pretrained on the general-domain corpus, which has a different data distribution from the domain corpora targeted to a certain domain. That is, directly adopting BERT models pretrained on a general corpus may underestimate the contribution of domain-specific knowledge hidden in the domain corpora, which is quite valuable for training DL models for IR tasks in the AEC domain (Zhang & El-Gohary, 2021). Unfortunately, to the best of our knowledge, there are no publicly available domain-specific corpora related to the AEC domain. No domain-specific BERT-based model is available for downstream IR tasks in the AEC domain. Consequently, it is still unclear how domain-specific corpora impact the performance of DL-based methods for IR tasks in the AEC domain.

To address this need, this work develops an open domain corpus and systematically investigates the performance of the various domain corpora and transfer learning strategies on deep learning models. Because the ARC processes involve almost all typical NLP and IR tasks in the AEC domain, this work takes the ARC tasks as typical cases. First, two transfer learning strategies are discussed in detail. Second, two domain corpora datasets were constructed. Third, various domain corpora and transfer learning strategies are investigated for various deep learning models and the downstream IR tasks (i.e., TC and NER). Finally, the RegulatoryBERT model is obtained utilizing the developed domain corpora based on the BERT model (Devlin et al., 2018), which achieved state-of-art results on all downstream IR tasks of ARC.

The remainder of this paper is organized as follows. Section 2 reviews the related work and highlights the potential research gaps. Section 3 describes the transfer learning strategies for AEC. Section 4 illustrates the dataset development. Sections 5 and 6 illustrate and analyze the results of the experiment. Section 7 discusses the advantages and contributions of this research and notes the limitations. Finally, Section 8 concludes this research.

## 2 Overview of related studies

### 2.1 Deep learning-based information retrieval methods in the AEC domain

Deep learning models are composed of multiple processing layers that enable the learning of data representations with multiple levels of abstraction and have shed light on sequential data such as text (LeCun et al., 2015). The main drawback of traditional machine learning models is the reliance on manual feature engineering, which is very time-consuming (Wu et al., 2022a). Deep learning models can extract features automatically from training text data. With the development of deep learning techniques and open datasets for model training, it is possible for deep NLP-based methods to achieve a more comprehensive understanding of regulatory texts (Fuchs, 2021). Deep learning approaches have outperformed traditional approaches in four major tasks in NLP, including classification, matching, translation, and sequence labeling (Li, 2017). Thus, with the advent of deep learning methods, IR methods for AEC have been further developed. Generally, the deep learning-based IR methods used in AEC can be divided into two tasks: (1) TC and (2) NER. Therefore, the existing researches on the TC and NER tasks are reviewed in the following two subsections (Sections 2.1.1 & 2.1.2). Moreover, deep learning models can also be divided into two main types: (1) BERT-based models and (2) traditional deep learning-based models (i.e., non-BERT models). The BERT-based model is a multilayer bidirectional transformer-decoder structure (Devlin et al. 2018) and is pretrained on large datasets. The structure and pretrained methods of BERT models differ from those of traditional models. Thus, the existing studies are categorized and summarized in Table 1, according to the tasks and models.

**2.1.1 Deep learning-based text classification**

The text classification (TC) task aims to recognize relevant sentences from large quantities of documents and assign them to one or more predefined categories (Manning & Schutze, 1999) to facilitate downstream tasks. Deep learning-based methods formulate the TC problem as a typical classification task, where each sentence/document is assigned a label. To deal with unstructured text data, the widely used deep learning-based methods first represent the sentences as vectors using word embedding techniques and then classify the vectors using deep learning models. Many deep learning models for TC, including TextCNN (Chen, 2015), TextRNN (Liu et al., 2016), TextRCNN (lai et al., 2015), DPCNN (Johnson et al., 2017), transformers (Vaswani et al., 2017), and BERT (Devlin et al., 2018), have been proposed in recent years. In the AEC domain, there has been a growing body of research efforts using deep learning-based methods to solve TC problems. For example, Tian et al. (Tian et al., 2021) classified on-site construction reports into six categories based on TextCNN. The result shows that the performance of TextCNN outperformed the traditional machine learning models. Cheng et al. (Cheng et al., 2020) deployed the gated recurrent unit (GRU) model for construction site accident classification. Zhong et al. (Zhong et al., 2020) developed a CNN model to classify accident narratives. Li et al. (Li et al., 2020) deployed a fastText-based classification model for analyzing construction accident claims. Fang et al. (Fang et al., 2020) classified near-miss information contained within safety reports using the BERT model, which achieved the best performance. To date, most of the studies have utilized traditional deep learning models for TC.

**2.1.2 Deep learning-based named entity recognition**

The named entity recognition (NER) task aims to identify and extract the structured information in the relevant sentences for downstream usages, including interpreting rules into computer-processable formats (Zhou et al., 2020; Zhang & El-Gohary, 2021) and construing knowledge graphs (Leng et al., 2019). Deep learning-based methods formulate the NER problem as a sequence labeling task, where each word or phrase is assigned a label (Zhang & El-Gohary, 2021). Similar to the methods for TC, the unstructured text data are also represented into vectors before model training and prediction. The typical deep learning models for NER include LSTM (Greff et al., 2017), BiLSTM-CRF (Huang et al., 2015), LSTM-CNN-CRF (Ma & Hovy, 2016), and BERT. In the AEC domain, there are now an increasing number of studies that focused on deep learning-based methods for the NER task. For example, the LSTM-based models are designed to identify semantic information elements for compliance checking purposes (Zhong et al., 2020; Moon et al., 2021; Zhang & El-Gohary, 2021; Feng & Chen, 2021; Moon et al., 2022) and for semiautomated knowledge graph construction (Leng et al., 2019). Zhou et al. (Zhou et al., 2020) utilized the pretrained BERT model for extracting predefined semantic labels in Chinese regulatory texts. Subsequentially, they formalized the extracted elements into language-independent pseudocodes. To date, most of the studies have utilized traditional deep learning models for NER.

Table 1 Deep learning-based information retrieval methods

| Tasks/Deep learning models | Traditional models | BERT-based models |
| --- | --- | --- |
| TC | Cheng et al., 2020; Zhong et al., 2020; Li et al., 2020; Tian et al., 2021 | Fang et al., 2020 |
| NER | Leng et al., 2019; Zhong et al., 2020; Zhang & El-Gohary, 2021; Moon et al., 2021; Feng & Chen, 2021; Moon et al., 2022; | Zhou et al., 2020 |

## 2.2 Transfer learning methods and domain corpora for AEC

Deep learning-based approaches require highly expensive manual effort to prepare enough training datasets (Xu & Cai, 2021). However, in the AEC domain, there are not sufficient training datasets or publicly available domain corpora for many domain-specific IR applications, such as TC and NER. To address this problem, various transfer learning strategies that can leverage labeled data from other domains have been proposed. Transfer learning is used to improve a model from one domain by transferring information from a related domain (Weiss et al., 2016). Deep transfer learning methods can be classified into four categories: instance-based methods, mapping-based methods, network-based methods, and adversarial-based methods (Tan et al., 2018). Instance-based methods utilize instances in the source domain by a specific weight adjustment strategy (e.g., TaskTrAdaBoost technology (Yao & Doretto, 2010)). Mapping-based methods map instances from the source domain and target domain into a new data space, where instances from two domains are similar (e.g., transfer component analysis (Pan et al., 2010)). Network-based methods, which are the most popular transfer techniques in NLP, refer to the reuse of the partial network pretrained in the source domain. Network-based methods can be mainly divided into two types, i.e., non-BERT-based models (named traditional deep learning methods) and BERT-based methods. Traditional deep learning methods and BERT-based models adopt different transfer learning methods. The two typical methods of this type are (1) pretraining transformer-based models (i.e., BERT-based models) on large corpora and then fine-tuning them for downstream text tasks (Fang et al., 2020; Zhou et al., 2020) and (2) pretraining word embedding models for traditional deep learning models on source-domain data (Zhang & El-Gohary, 2021). Adversarial-based methods use adversarial technology to find transferable features that are both suitable for two domains (Tan et al., 2018).

**2.3 Research gaps**

In the AEC domain, some research efforts have explored the use of deep learning-based models in AEC downstream IR tasks (i.e., TC and NER). Despite the importance of these efforts, there are two knowledge gaps that this paper aims to address. First, the existing studies focused on only training or fine-tuning deep learning models on their own application-specific datasets. Few studies have explored transfer learning techniques to overcome the challenge of lacking labeled training data. Second, there are few open publicly available domain corpora prepared for transfer learning techniques in the AEC domain, which hampers the use of transfer learning methods. In conclusion, there is no systemic discussion of how to improve the performance of deep learning-based models utilizing domain corpora and transfer learning techniques, which require less human effort than labeling training datasets.

To address these gaps, this work aims to handle the following four scientific issues: (1) constructing AEC domain corpora to facilitate transfer learning usages, (2) exploring whether domain corpora are more efficient than general corpora for enhancing model performance, (3) investigating the value of domain corpora with various data distributions, and (4) investigating the performances of various transfer learning strategies for widely used deep learning models in the AEC domain.

# 3 Methodology

To improve the performance of deep learning models on IR tasks such as TC and NER without increasing the effort of manual annotation, we systematically illustrate and analyze two typical domain corpus-based transfer learning methods for two types of deep learning models (i.e., BERT-based models and traditional deep learning methods). The two widely used transfer learning techniques are introduced in Sections 3.1 and 3.2.

The investigation workflow of various domain corpus-enhanced transfer learning methods is shown in Fig. 1. The proposed workflow consists of three parts: (1) domain corpora development, where two domain corpora are constructed (Section 4.1), (2) transfer learning techniques deployment, where two transfer learning techniques (i.e., word embedding models and BERT-based further pretraining techniques) are deployed based on the constructed domain corpora (Section 4.2), and (3) deep learning models development and evaluation, where several deep models are developed and evaluated for TC and NER tasks (Section 5). The core toolkits used in the three parts are also listed at the bottom in Fig. 1. The workflow is conducted based on the Python script.

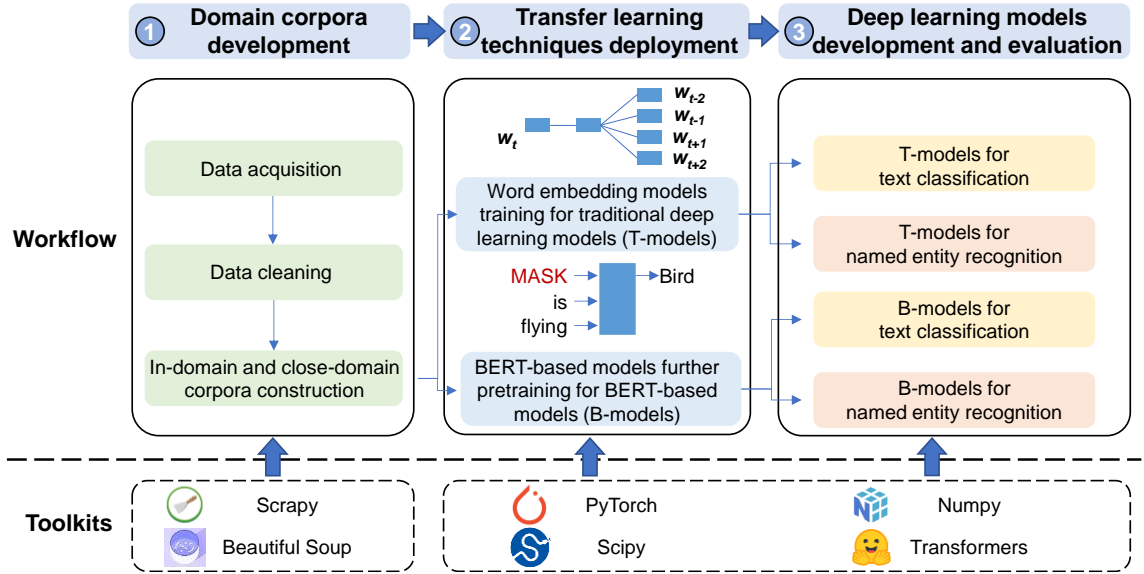

Fig. 1 The investigation workflow of various domain corpus-enhanced transfer learning methods

### 3.1 Domain corpus enhanced for traditional deep learning methods

The basic training and prediction flow of traditional deep learning models is shown by the blue arrow in Fig. 2. The flow consists of two main steps: (1) a word embedding model, which represents each word in a text into a low-dimensional real number vector, thus obtaining the text's vector representation, and (2) a deep learning model, which completes the downstream tasks (i.e., TC or NER) based on the text's vector representation. For the existing studies in the AEC domain, the word embedding models are either random initial (Tian et al., 2021) or pretrained on general corpora (Zhang & El-Gohary, 2021). The general corpora usually consist of the internet news corpora and the encyclopedia corpora (e.g., Wikipedia corpus (Wikipedia, 2021a) and Sogou news (Sogou, 2021)), where AEC-specific domain-related texts are rare.

The flow of domain corpus-based transfer learning methods for traditional deep learning models is shown by the red arrow in Fig. 2. The flow consists of two main steps: (1) AEC domain-specific word embedding models, which are pretrained on domain corpora that contain more domain-specific terms, and (2) word embedding model replacement, where the word embedding models pretrained on the general corpora are ousted by the AEC domain-specific model. Some research efforts have shown that domain-specific word embedding models can capture domain-specific terms better than general terms (Wang et al., 2018). Therefore, domain-specific word embedding models may also be useful for downstream tasks. It is important to note that the word embedding pretraining task is a typically unsupervised task and does not require additional manual labeling.

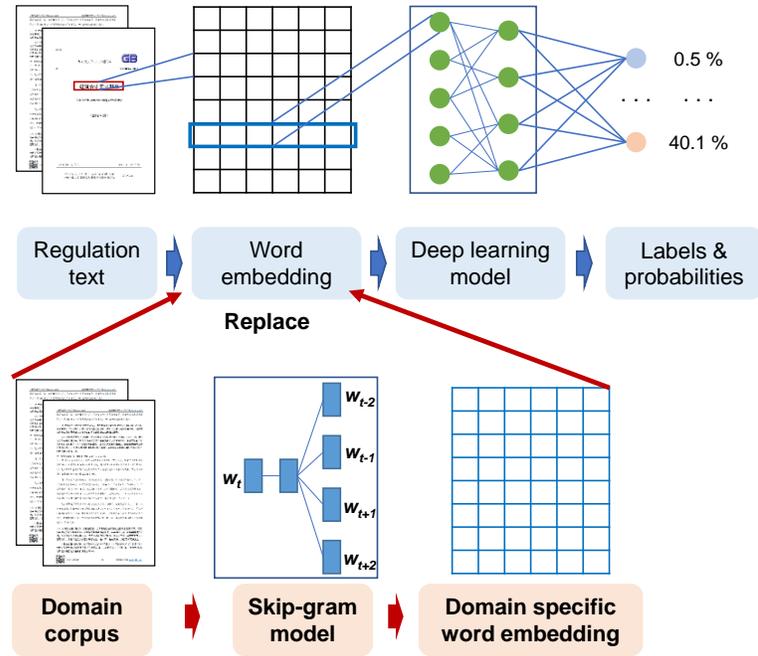

Fig. 2 Domain corpus-enhanced traditional deep learning methods

## 3.2 Domain corpus enhanced for BERT-based methods

The basic training and prediction flow of the BERT-based model is shown by the blue arrow in Fig. 3. The flow consists of two main steps: (1) word embedding and encoding, which first embeds all tokens of the input sentence and then uses a pretrained BERT model to encode the input embeddings to contextual representations, and (2) fine-tuning and prediction, where the contextual representations are then input to the fine-tuning layers to obtain the prediction result. For the downstream tasks, only the parameters of the last few layers (i.e., the fine-tuned layers) need to be optimized during model training, also known as fine-tuning. Therefore, the BERT-based model can achieve outstanding performance when the training datasets are relatively small.

However, the widely used BERT-based models are pretrained in the general-domain corpus (Sun et al., 2019), which has a different data distribution from that of the target domain. When the training datasets are very limited, the BERT-based models pretrained in the general-domain corpus still struggle to obtain satisfactory results.

A natural idea is to further pretrain BERT with the target domain data. The flow of domain corpus-based transfer learning methods for BERT-based models is shown by the red arrows in Fig. 3. The flow consists of two main steps: (1) AEC domain-specific BERT-based model pretraining and (2) pretrained BERT model replacement. First, we can further pretrain BERT on domain-specific data with a masked language model and a next-sentence prediction task (Devlin et al., 2018). Then, the AEC domain-specific BERT models can be used to replace the BERT model pretrained in the general domain. After the replacement, the fine-tuning and prediction tasks are the same as before mentioned. It is important to note that the masked language model and the next-sentence prediction task are typically unsupervised tasks that do not require additional manual labeling.

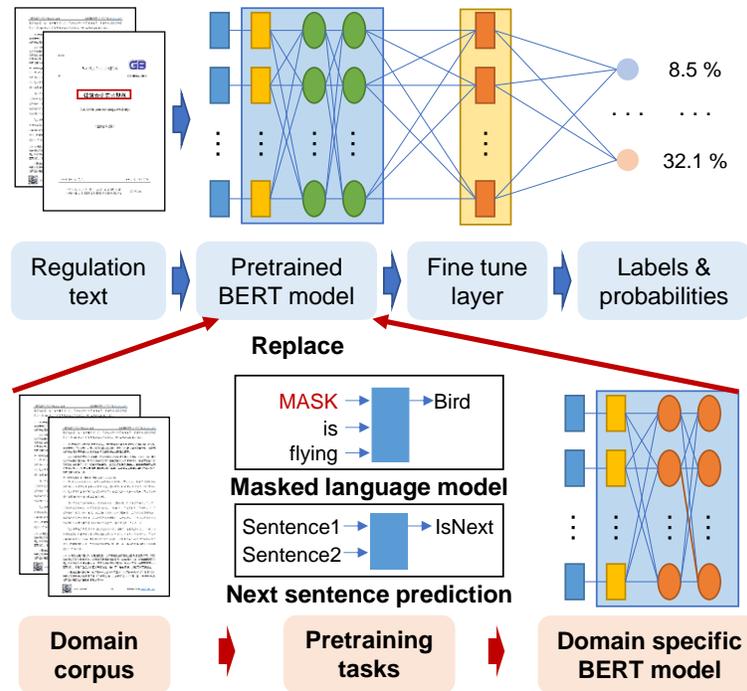

Fig. 3 Domain corpus-enhanced BERT-based methods

## 4 Domain corpora development and pretraining configuration

### 4.1 Domain corpora development

To analyze the performance of transfer learning techniques utilizing various domain corpora, this work collects a large number of domain corpora and subsequently constructs in-domain corpora and close-domain corpora, respectively. The civil engineering regulatory texts are the object of analysis for the ARC downstream tasks. Therefore, the in-domain corpora mainly contain regulatory texts. The close-domain corpora contain the AEC domain text, such as the definition of various terms in the AEC domain. The construction processes of the in-domain corpus and close-domain corpus are described in Sections 4.1.1 and 4.1.2, respectively.

The statistical information of the constructed corpora is summarized in Table 2. Note that although the number of lines in the close-domain corpus is smaller than that of the in-domain corpus, the number of texts in the close-domain corpus is close to that of the in-domain corpus. Because in the close-domain corpus, one line contains more texts. The developed domain corpora can be found in https://github.com/SkydustZ/AEC-domain-corpora/tree/main/domain%20corpus.

### 4.1.1 In-domain corpora construction

The data distribution of the in-domain data is the same as the training data distribution. In this research, the training and validation datasets (Section 5.1) are from Chinese regulatory text; thus, the in-domain corpora consist of the Chinese regulatory texts crawled from the *website of the Chinese codes* (Soujianzhu, 2021). The construction process consists of (1) data acquisition and (2) data cleaning. First, a total of 396 design codes are crawled based on scrapy (Myers & McGuffee, 2015). The obsolete design codes are also taken into account to build a sufficiently large in-domain corpora. Data cleaning was then performed, and the cleaned in-domain corpus contained 126,433 lines and 10,895,634 Chinese characters.

**4.1.2 Close-domain corpora construction**

The data distribution of close-domain data is close to the training data distribution. In this research, the close-domain corpora consist of texts from the pages of the civil engineering category in Wikipedia (Wikipedia, 2021b) and texts from the civil engineering section in *Encyclopedia of China* (Encyclopedia of China Publishing House, 2009).

For the Wikipedia corpora, the construction process consists of (1) data acquisition and (2) data cleaning. The pages and subpages of the civil engineering category in Wikipedia contain various civil engineering-related terms and their corresponding definitions. First, a total of 11,189 terms are crawled from the pages based on scrapy (Myers & McGuffee, 2015). Afterward, data cleaning was performed to sift the texts that were not relevant to the AEC domain. After data cleaning, a total of 10,488 terms and their corresponding descriptions were collected. The definition of one concept is stored in one line; thus, 10,488 lines are collected.

The *Encyclopedia of China* corpus is obtained by parsing the text and performing data cleaning. Then, a total of 16,238 lines of texts are obtained. The above two are aggregated together to form the close-domain corpus, which contains 26,727 lines and 12,899,562 Chinese characters in total.

Table 2 Statistical information of the constructed domain corpora

|  | Number of lines | Number of Chinese characters |
| --- | --- | --- |
| In-domain corpus | 126,433 | 10,895,634 |
| Close-domain corpus | 26,727 | 12,899,562 |

## 4.2 Pretraining configuration

**4.2.1 Pretrained word embedding models for traditional deep learning models**

For word embedding model training, the Wikipedia Chinese corpora (Wikipedia, 2021a) are used together with the domain corpora, as the domain corpora are too small to contain all common words. Therefore, three word embedding models are trained: (1) the general model, which is trained only using the Wikipedia Chinese corpora and is set as the control group; (2) the in-domain model, which is trained using the Wikipedia Chinese corpora and the in-domain corpora; and (3) the close-domain model, which is trained using the Wikipedia Chinese corpora and the close domain corpora.

To train the word embedding models, Chinese word segmentation is performed on the above corpora using the jieba toolkit. The jieba toolkit is an open-source Chinese segmentation application in Python language. Subsequently, the segmented corpora are used to train the word embedding model utilizing the skip-gram method (Mikolov et al., 2013) with the negative-sampling technique (Mikolov et al., 2013). The skip-gram method uses the headword to predict the context words around it with a similar basic theory. The negative-sampling technique is a more efficient way of deriving word embeddings. The word embedding models are trained with a dimension of 300, which is widely used by many researchers (Li et al., 2018; Wang et al., 2018) and with a negative-sampling number of 5.

**4.2.2 Further pretraining the BERT-based models**

For the further pretraining of the BERT-based models, two domain corpora are considered: (1) the in-domain corpora and (2) the close-domain corpora. The control group is the model without further pretraining, i.e., pretrained on only the general corpora.

In addition, we investigated the further pretraining performance of various BERT-based models. Two BERT-based models are chosen: (1) bert-base-chinese (Hugging Face, 2019) and (2) ERNIE (Sun et al., 2019). The bert-base-chinese model, which is pretrained on the Wikipedia Chinese corpora, is the most common BERT-based model for tasks in Chinese. The ERNIE model was published by Baidu and is pretrained based on the Baidu Chinese corpora and entity-level masking and phrase-level masking techniques, and it performs better than the BERT model on some downstream tasks (Sun et al., 2019). The models are further pretrained with a learning rate of $5 \times 10^{-5}$ and a batch size of 4. The RegulatoryBERT and RegulatoryERNIE models were pretrained with in-domain corpora. The CivilBERT and CivilERNIE models were pretrained using close-domain corpora.

## 5 Experiments and analysis

### 5.1 Datasets

We evaluate our approach and domain corpora on two datasets, including the regulatory TC dataset and NER dataset. The two datasets are briefly introduced as follows.

#### 5.1.1 Regulatory text classification dataset

The regulatory TC dataset developed by the authors (Zheng et al., 2022) is used in this work. Seven text categories, including direct, indirect, method, reference, general, term, and others, were defined by the authors based on the computability of a text. The dataset distribution is shown in Fig. 4.

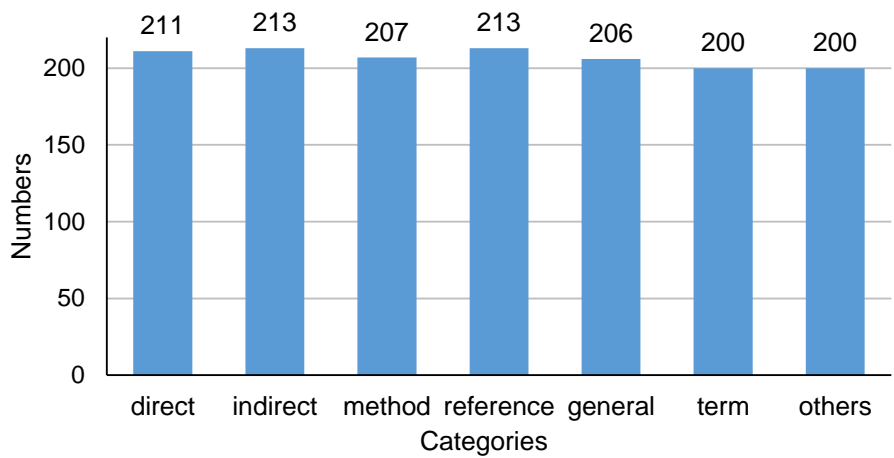

Fig. 4 Distribution of the text classification dataset

#### 5.1.2 Named entity recognition dataset

The NER dataset developed by Zhou et al. (Zhou et al., 2020) is used in this work. Seven semantic labels, including obj, sobj, prop, cmp, Rprop, ARprop, and Robj, were defined by the authors (Zhou et al., 2020) based on the hierarchical structure of the objects and attributes in the BIM model. Then, the gold standard for semantic labeling was developed (Zhou et al., 2020). The dataset includes 611 sentences and has a total number of 4336 semantic elements. A repository containing the dataset was established on GitHub at https://github.com/Zhou-Yucheng/auto-rule-transform. The dataset distribution is shown in Fig. 5.

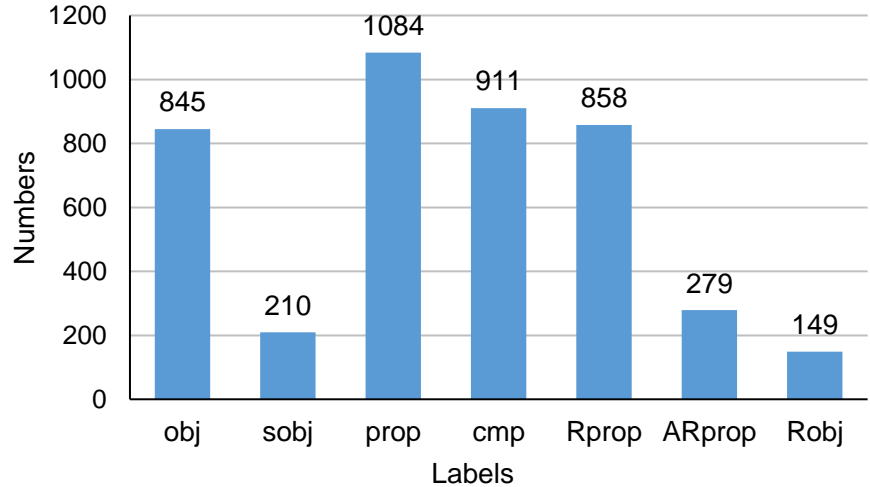

Fig. 5 Distribution of the named entity recognition dataset

## 5.2 Evaluation performance metrics

To measure the result, the model predictions are compared with the gold standard, and the widely used weighted average F1 score (weighted F1) is selected.

First, the precision (P), recall (R), and F1 score (F1) are calculated for each semantic label:

$$P = N_{correct}/N_{labeled} \tag{1}$$

$$R = N_{correct}/N_{true} \tag{2}$$

$$F_1 = 2PR/(P + R) \tag{3}$$

where $N_{\{correct, labeled, true\}}$ denotes the number of {model correctly labeled, model labeled, true} elements for a label. Finally, the weighted average F1 score is calculated to represent the overall performance ($n_i$ denotes the number of elements of the i-th semantic label):

$$\text{Weighted } F_1 = \left(\sum_i n_i F_{1,i}\right)/\sum_i n_i \tag{4}$$

## 5.3 Experiments

To investigate the various domain corpora and transfer learning techniques for the IR tasks (i.e., TC & NER) in the ARC domain, several experiments are conducted in this section. The two datasets are randomly split into training and validation datasets at a 0.8:0.2 ratio, where the training dataset is used to train and update the DNN model, and the validation dataset is used to test the performance of the model.

### 5.3.1 Experiment 1: domain corpus-enhanced word embedding for TC

This experiment aims to investigate the performance of domain-specific pretrained word embeddings on TC tasks. The three pretrained word embedding models in Section 4.2.1 are combined with various deep learning-based TC models. Six types of deep learning-based TC models are employed to perform the experiment: (1) TextCNN (Chen, 2015), (2) TextRNN (Liu et al., 2016), (3) TextRNN with attention (TextRNN-Att), (4) TextRCNN (lai et al., 2015), (5) DPCNN (Johnson et al., 2017), and (6) Transformers (Vaswani et al., 2017). The above models are trained with an epoch of 100 and a padding size of 64. In addition, the effects of various learning rates are considered via grid search on learning rates of 0.001, 0.0005, 0.00025, and 0.0001. The highest metrics in the validation datasets are chosen as the optimal metrics. The training results of the models are shown in Table 3.

Table 3 shows that in the TC task, there are slight improvements in most of the deep models except the TextRCNN model using domain-specific word embedding models instead of the general one. The word embedding models trained on the in-domain corpus are suitable for RNN-based models (i.e., TextRNN and TextRNN-Att). The word embedding models trained on the close-domain corpus are suitable for the transformer model. However, there is no word embedding model that works for all models. This result is in line with the findings of Wang et al. (Wang et al., 2018) in the biomedical informatics domain. There is no consistent global ranking of word embeddings for TC applications in the ARC domain. However, considering the improved performance of most deep models, the use of domain-specific word embeddings could be considered when employing deep models for TC in the ARC domain. Especially when employing the RNN-based model, word embedding models trained on an in-domain corpus could be considered.

Table 3 Weighted F1 score on the text classification datasets

| Model/Word embedding | General | In-domain | Close-domain |
|---|---|---|---|
| TextCNN | 86.26% | 85.60% | **86.72%** |
| TextRNN | 72.16% | **76.69%** | 72.29% |
| TextRNN-Att | 81.50% | **81.64%** | 79.32% |
| TextRCNN | **86.63%** | 84.72% | 83.93% |
| DPCNN | 84.52% | **86.56%** | 82.66% |
| Transformers | 73.96% | 73.87% | **75.05%** |

**5.3.2 Experiment 2: domain corpus-enhanced word embedding for NER**

This experiment aims to investigate the performance of domain-specific pretrained word embeddings on NER tasks. The three pretrained word embedding models in Section 4.2.1 are combined with various deep learning-based NER models. Four types of deep learning-based NER models are employed to perform the experiment: (1) LSTM (Greff et al., 2017), (2) BiLSTM, (3) BiLSTM-CRF (Huang et al., 2015), and (4) BiLSTM-CNN-CRF (Ma & Hovy, 2016). The models are trained with an epoch of 1000 and a batch size of 20. In addition, the effects of various learning rates are considered via grid search on learning rates of 0.015, 0.01, 0.005, and 0.001. The highest metrics in the validation dataset are chosen as the optimal metrics. The training results of the models are shown in Table 4.

Table 4 shows that in the ARC domain NER task, there are slight negative effects on all the deployed deep models when using domain-specific word embedding models instead of using the general one. Therefore, when using traditional deep models for ARC NER tasks, it is not necessary to train domain-specific word embedding models. The use of an easily accessible general word embedding model may be sufficient.

Table 4 Weighted F1 score on the named-entity recognition datasets

| Model/Word embedding | General | In-domain | Close-domain |
| --- | --- | --- | --- |
| LSTM | **73.62%** | 72.93% | 73.31% |
| BiLSTM | **73.63%** | 72.93% | 72.58% |
| BiLSTM-CRF | **76.70%** | 76.15% | 76.54% |
| BiLSTM-CNN-CRF | **75.72%** | 75.19% | 74.74% |

### 5.3.3 Experiment 3: further pretrained BERT model for TC

This experiment aims to investigate the performance of further pretrained BERT-based models for TC tasks. Four further pretrained BERT-based models in Section 4.2.2 are applied to the TC task. In addition, the effects of various learning rates are discussed. The BERT-based models are fine-tuned on the training datasets with an epoch of 100 and a padding size of 64. The highest metrics in the validation dataset are chosen as the optimal metrics. The results are shown in Table 5.

Table 5 shows that in the ARC domain TC task, almost all further pretrained models on in-domain corpora (i.e., RegulatoryBERT and RegulatoryERNIE) perform better than the original BERT-based models (i.e., BERT and ERNIE). Among them, RegulatoryBERT achieved the global best weighted F1 score of 92.12%. However, the performances of the further pretrained models on the close-domain corpora (i.e., CivilBERT and CivilERNIE) are not good enough. In some cases, the performances of CivilBERT and CivilERNIE are even worse than those of the original model. Generally, further pretraining on in-domain corpora can bring better performance than further pretraining on close-domain corpora. This experiment illustrates that further pretraining on in-domain corpora is very useful to improve the performance of BERT for the TC task.

Table 5 Weighted F1 score on the text classification datasets

| Lr/model | BERT | RegulatoryBERT | CivilBERT | ERNIE | RegulatoryERNIE | CivilERNIE |
| --- | --- | --- | --- | --- | --- | --- |
| 1e-5 | 89.58% | **90.53%** | 88.08% | 87.29% | **91.25%** | 88.77% |
| 3e-5 | 88.91% | **90.55%** | 89.72% | **91.37%** | 89.70% | 87.31% |
| 5e-5 | 89.80% | **92.12%** | 89.72% | 89.64% | **90.43%** | 86.24% |
| 7e-5 | 86.47% | 88.86% | **89.00%** | 86.41% | **90.46%** | 88.68% |

### 5.3.4 Experiment 4: further pretrained BERT model for NER

This experiment aims to investigate the performance of further pretrained BERT-based models for NER tasks. Four further pretrained BERT-based models in Section 4.2.2 are applied to the ARC NER task. In addition, the effects of various learning rates are discussed. The BERT-based models are fine-tuned on the training datasets with an epoch of 30 and a batch size of 16. The highest metrics of the validation datasets are chosen as the optimal metrics. The results are shown in Table 6.

Table 6 shows that in the ARC domain NER task, the further pretrained BERT model on in-domain corpora (i.e., RegulatoryBERT) performs better than the original BERT-base model. There are slight improvements under all learning rates when using RegulatoryBERT instead of the original one. Moreover, RegulatoryBERT achieves the global best weighted F1 score of 86.8%, which is better than the result of Zhou et al. (Zhou et al., 2020). However, other further pretrained models (i.e., CivilBERT, CivilERNIE, and RegulatoryERNIE) do not perform well enough. In some cases, the performances are even worse than those of the original model. This experiment illustrates that the further pretrained RegulatoryBERT model is useful for improving the performance of the ARC NER task.

Table 6 Weighted F1 score on the named-entity recognition datasets

| Lr/model | BERT | RegulatoryBERT | CivilBERT | ERNIE | RegulatoryERNIE | CivilERNIE |
|---|---|---|---|---|---|---|
| 1e-5 | 86.4% | **86.7%** | 86.0% | 77.2% | **78.9%** | 77.7% |
| 3e-5 | 86.4% | **86.8%** | 85.3% | **79.7%** | 76.7% | 78.1% |
| 5e-5 | 86.2% | **86.3%** | 86.0% | **81.0%** | 80.5% | 80.5% |
| 7e-5 | 86.3% | **86.5%** | 85.9% | 78.5% | 79.7% | **80.4%** |

## 5.4 Summary
### 5.4.1 Text classification task

For the TC task in the ARC domain, the performances of various models on the validation dataset are shown in Fig. 6. Almost all models except the TextRCNN model are enhanced by the domain corpus and transfer learning techniques. Among all models, the TextRNN model achieves the most notable performance improvement, which increases the weighted F1 score by 4.5%. The second is the further pretrained BERT-based model (i.e., the RegulatoryBERT model), which increases the weighted F1 score by 2.3%. Meanwhile, the RegulatoryBERT model achieves the highest performance among all models, with a weighted F1 score of 92.1%. Therefore, for the TC task in the ARC domain, the domain corpus can be used to train domain-specific word embedding models or further pretrain the BERT model to improve the performance.

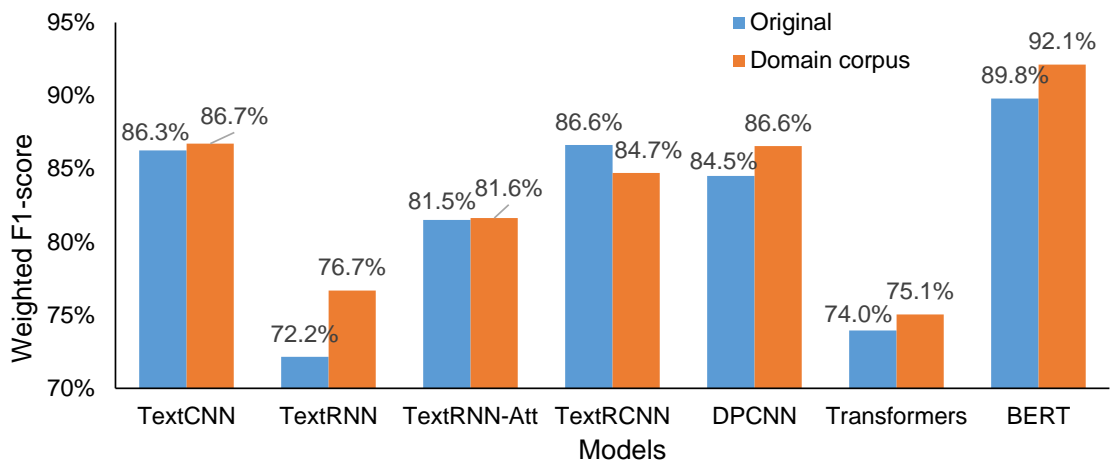

Fig. 6 Performance of various models on the TC task

### 5.4.2 Named entity recognition task

For the NER task in the ARC domain, the performances of the various models on the validation dataset are shown in Fig. 7. Unlike the TC task, there are slight negative effects on all the deployed deep models when using domain-specific word embedding models instead of using the general one. The only performance improvement is obtained by the further pretrained BERT-based model (i.e., the RegulatoryBERT model). The RegulatoryBERT model achieves a slight performance improvement, which increases the weighted F1 score by 0.4%. Meanwhile, the RegulatoryBERT model achieves the highest performance among all models, with a weighted F1 score of 86.8%. Therefore, for the NER tasks in ARC, it is not necessary to train domain-specific word embedding models. Instead, further pretraining of BERT-based models on a domain corpus is still effective for improving performance.

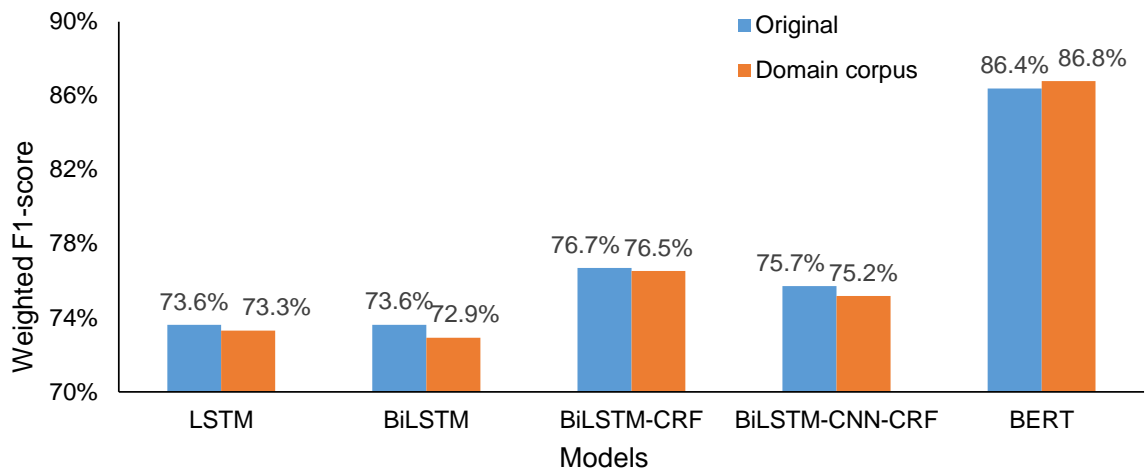

Fig. 7 Performance of various models on the NER task

### 5.4.3 Theoretical explanation

In general, BERT-based models perform better than traditional models. A possible reason is that BERT-based models could consider context-related information hidden in the text. Traditional models use static word embedding, where each word has a single vector, regardless of context. All senses of a polysemous word share the same representation. While utilizing the masked language model, the BERT-based model can create contextualized word representations, where the word vectors are sensitive to the context in which they appear (Ethayarajh, 2019). Therefore, using contextualized word representations better allows the machine to obtain a full understanding of the text.

Meanwhile, in the TC task, the best BERT-based model (i.e., RegulatoryBERT) improved the weighted F1 score by 5.4% over the best traditional deep learning model (i.e., TextCNN-enhanced). For the NER task, the best BERT-based model (i.e., RegulatoryBERT) improved the weighted F1 score by 10.1% over the best traditional deep learning model (i.e., BiLSTM-CRF). The BERT-based model improves more significantly on the NER task than on the TC task. This may be because the TC task is a sentence-level classification task and is less sensitive to context-related information in a sentence. The sentence embedding is the average of the word embeddings, which is not as sensitive to the embedding vector of a single word. In contrast, the NER task is a word-level sequence labeling task and is more sensitive to context-related information in sentences. Therefore, the contextualized word representations provided by the BERT-based model are more useful to the NER task.

## 6 Discussion

This work offers a leading initiative by developing the first domain corpora and enhancing DL-based transfer learning techniques for various IR tasks in the AEC domain. To improve the performance of deep learning models on ARC downstream tasks without increasing the effort of manual annotation, we systematically illustrate and analyze two typical domain corpus-based transfer learning methods for traditional deep learning models and BERT-based models. In comparison to the previous effort, this work contributes to the body of knowledge on four main levels:

(1) Word embedding models and further pretraining techniques are typical unsupervised tasks that can enhance the models with unlabeled domain corpora and eliminate extensive labeling efforts. To systematically analyze the effect of domain corpora and transfer learning techniques on ARC downstream tasks, in-domain corpora and close-domain corpora are developed in this work. To the best of our knowledge, this is the first publicly available domain corpora for transfer learning in the AEC domain.

(2) Based on the proposed domain corpora, several domain-specific word embedding models (i.e., in-domain model and close-domain model) and further pretrained BERT-based models (i.e., RegulatoryBERT, CivilBERT, RegulatoryERNIE, and CivilERNIE) are obtained.

(3) Six types of deep learning-based TC models and four types of deep learning-based NER models are deployed for the experiments. Subsequently, four experiments are conducted to reach the following conclusions. 1) There is no consistent global ranking of word embeddings for the TC task. However, the domain-specific word embedding models can slightly prompt the performance of most deep models for TC tasks in the ARC domain. 2) For the NER tasks, it is not necessary to train domain-specific word embedding models. The use of an easily accessible general word embedding model may be sufficient. 3) Further pretraining on an in-domain corpus is very useful for improving the performance of BERT for the TC task. 4) The further pretrained RegulatoryBERT model is useful for improving the performance of the NER tasks in the ARC domain. 5) The BERT-based models perform better than the traditional models on both TC tasks and NER tasks. Compared with the traditional models, the BERT-based model improves more significantly on the NER task than the TC task by considering context-related information hidden in sentences; detailed explanations are discussed in Section 5.4.

(4) The RegulatoryBERT is further pretrained based on the proposed in-domain corpora and achieves better performance than the original models (i.e., BERT & ERNIE) for all downstream IR tasks (i.e., TC and NER) in ARC. This study also open-sources the model, which can help future related studies in ARC.

In the AEC domain, one challenge of deep learning models is the lack of labeled training data. The results show that the developed domain corpora, associated with transfer learning techniques, can enhance the performance of deep learning models without extra manual annotation. This can facilitate the development and further application of deep learning models in the ARC and AEC domains. These deep learning models can be used for TC and NER tasks, which are the two most vital tasks for the ARC domain. Typical further applications may include 1) automated rule interpretation by which researchers can filter irrelevant texts by TC to save time and improve accuracy for the subsequent tasks (Zhou & El-Gohary, 2016); 2) assigning more proper code generation functions to various kinds of regulatory sentences by identifying the features by TC, which aims to improve the coverage of automated rule interpretation (Solihin & Eastman, 2015); 3) using TC in construction site accident report classification (Cheng et al., 2020) and topic modeling (Lin et al., 2020); 4) using NER tasks to automatically extract information from texts to serve the automated interpretation application or to serve the recommender system for relevant rule clauses (Zhou et al., 2020; Moon et al., 2022); and 5) using NER for semiautomatic construction and extensions of knowledge graphs, which are currently mainly approached via manual methods (Leng et al., 2019).

Despite the success of our approaches, many future studies would be valuable for both academia and industry. For example, the performance of the proposed RegulatoryBERT model should be further validated on other datasets, and more detailed parameter analysis on the influence of word embedding dimensions or model types (e.g., skip-gram model or CBOW) should be carried out in the future. Furthermore, the domain corpora can be further completed by containing more related texts to prompt the development of the AEC domain. In addition, despite the word embedding and further pretraining techniques analyzed in this work, more transfer learning and semi-supervised learning strategies should be explored for leveraging large-scale, pattern-rich general-domain corpora to solve downstream tasks in the AEC domain.

## 7 Conclusion

In this research, we systematically investigate how the domain corpus could enhance traditional deep learning models and BERT-based models for various IR tasks in the AEC domain. First, both in-domain and close-domain corpora are developed and opened for further exploration and adoption. To the best of our knowledge, this is the first publicly available domain corpora for pretraining NLP models in the AEC domain. Then, based on various experiments, we illustrate the advantages of developed domain corpora and BERT-based models and attempt to explain possible reasons behind these results:

1) For the TC tasks, domain-specific corpora can enhance both traditional word embedding models and BERT-based models, achieving 4.5% and 2.3% improvements in the F1 score, respectively.

2) For the NER tasks, domain corpora have a negative effect on traditional word embedding models and could slightly improve the performance of BERT-based models.

3) For all the tested IR tasks, BERT-based models such as RegulatoryBERT dramatically outperform traditional word embedding models, with maximum improvements of 5.4% and 10.1% in F1 score for TC and NER tasks, respectively.

4) Compared to traditional word embedding models, BERT-based models can capture the rich context-related information of a word or phrase in a sentence; thus, their performance in NER tasks is significantly better than that of traditional word embedding methods. Even in sentence-level tasks such as TC, BERT-based models perform better than traditional models.

Finally, a pretrained BERT model called RegulatoryBERT is proposed and obtains a 92.1% overall F1 score for the TC task and an 86.8% overall F1 score for the NER task. The proposed model achieves state-of-the-art performance on all AEC downstream IR tasks. The developed domain corpora and the RegulatoryBERT model demonstrate promising results in various IR tasks and may shed light on various future investigations and applications in the AEC domain.

## Acknowledgment

The authors are grateful for the financial support received from the National Natural Science Foundation of China (No. 51908323, No. 72091512), the National Key R&D Program (No. 2019YFE0112800), and the Tencent Foundation through the XPLORER PRIZE.